\documentclass[runningheads]{llncs}
\usepackage{graphicx}
\DeclareGraphicsExtensions{.pdf,.PDF}

\usepackage{booktabs}
\usepackage{siunitx}
\usepackage{multirow}
\usepackage{amsmath}
\usepackage{algorithm,algorithmic}
\usepackage[capitalise]{cleveref}

\DeclareMathOperator*{\argsort}{\arg\!sort}

\usepackage{tabularx}
\usepackage{subcaption}
\begin{document}
\title{Mining Feature Relationships in Data}
%
%
\author{Andrew Lensen\orcidID{0000-0003-1269-4751}}
\institute{School of Engineering and Computer Science,\\
    Victoria University of Wellington, PO Box 600, Wellington 6140, New Zealand\\
    \email{andrew.lensen@ecs.vuw.ac.nz}}

\maketitle              
\begin{abstract}
When faced with a new dataset, most practitioners begin by performing exploratory data analysis to discover interesting patterns and characteristics within data. Techniques such as association rule mining are commonly applied to uncover relationships between features (attributes) of the data. However, association rules are primarily designed for use on binary or categorical data, due to their use of rule-based machine learning. A large proportion of real-world data is continuous in nature, and discretisation of such data leads to inaccurate and less informative association rules. In this paper, we propose an alternative approach called feature relationship mining (FRM), which uses a genetic programming approach to automatically discover symbolic relationships between continuous or categorical features in data. To the best of our knowledge, our proposed approach is the first such symbolic approach with the goal of explicitly discovering relationships between features. Empirical testing on a variety of real-world datasets shows the proposed method is able to find high-quality, simple feature relationships which can be easily interpreted and which provide clear and non-trivial insight into data.

\keywords{Association Rule Mining \and Feature Relationships \and Feature Construction \and Feature Analysis \and Unsupervised Learning}
\end{abstract}

\section{Introduction}
Exploratory data analysis (EDA) is a fundamental task in the data mining process, in which data scientists analyse the properties and characteristics of different features (or instances) in a dataset, and the relationships between them \cite{tukey1977exploratory}. Simple linear feature relationships can be discovered through the use of statistical techniques such as Pearson's or Spearman's correlations.

Non-linear relationships are generally found by performing association rule mining (ARM) \cite{agrawal1993mining}, a rule-based machine learning method that produces rules that represent relationships between discrete features in a dataset. In the case of continuous data, discretisation techniques are commonly applied before performing ARM, limiting the quality and increasing the complexity of rules. 

Genetic programming (GP) is, perhaps, most known for its success in symbolic regression tasks: the canonical tree-based GP is intrinsically suited to representing non-linear regression models. The use of GP for \textit{interpretable} symbolic regression --- where a user can understand the operation of the evolved function --- has also been very successful \cite{dick2014bloat}. 

The above properties make GP a natural choice for discovering interpretable relationships between continuous variables in data. However, no such approach has yet been proposed; all existing uses of GP for ARM use either a rule-based grammar, or discretise the input space. 

This paper aims to propose the first approach to mining feature relationships (FRs), which are \textit{symbolic} representations of intrinsic relationships between features in a dataset. A new GP method will be developed which uses a fitness function that considers both the quality of the discovered FR, as well as the potential interpretability of the FR. A speciation-based approach will also be proposed to allow for multiple distinct and complementary FRs to be automatically found as part of a single evolutionary search.

\section{Background}
GP has seen significant success in recent years in feature analysis applications. Tree-based GP \cite{poli2008field}, in particular, has been widely used due to its functional structure, which is well-suited to mapping a set of input features to a new \textit{constructed feature} \cite{neshatian2012filter,tran2016genetic,hart2017hybrid}. 

The use of GP for feature construction for regression and unsupervised learning tasks are perhaps the most closely related areas to this work: evolving feature relationships can be seen as a form of ``unsupervised regression''. Several works have suggested the use of methods to limit model complexity in symbolic regression, either to improve interpretability or generalisability. These include parsimony pressure and other bloat control strategies \cite{dick2014bloat} as well as complexity measures such as Rademacher complexity \cite{chen2020rademacher}. The discovery and combination of ``subexpressions'' in GP (i.e.\ feature construction) was also shown to improve performance on symbolic regression tasks \cite{arnaldo2014multiple}. GP has been used for unsupervised tasks such as clustering \cite{handl2007evolutionary} and nonlinear dimensionality reduction \cite{mcDermott2019why,lensen2019multi}, often with a focus on interpretability \cite{lensen2020genetic}.

\subsection{Related Work}
A number of evolutionary computation approaches to ARM have been proposed \cite{telikani2020survey}, with most using a vector-based representation such as a genetic algorithm (GA) \cite{martin2014new} or particle swarm optimisation (PSO) \cite{kuo2011application}. The small number of papers using GP for ARM can be categorised into two paradigms: those using Genetic Network Programming (GNP) \cite{taboada2007assocation,mabu2011intrusion}, and those using a grammar-based G3P approach \cite{mara2012design,luna2018mining}. Of these, only a handful address the task of mining ARMs from continuous data \cite{telikani2020survey} (known as quantitative association rule (QAR) mining). These QAR methods, however, are all still constrained by the use of a grammar or network programming structure, and so they are unable to represent the relationships between continuous features in a more intuitive and precise symbolic manner.

\newpage
\section{Proposed Method: GP-FRM}
The proposed method, Genetic Programming for Feature Relationship Mining (GP-FRM), aims to evolve compact rules (trees) that reconstruct a feature of the dataset from other features. In this way, the learnt rule represents a relationship between a given (``target'') feature and a set of other features. A simple example is the tree $f_2 = f_1 \times f_0$, which is a non-linear relationship that would not be discovered by association rule mining algorithms. Such relationships are common: for example, the Body Mass Index (BMI) is a well-known "target" feature in the medical domain which is based on a person's mass ($m$) and height ($h$): $BMI = \frac{m}{h^2}$. As GP-FRM is an unsupervised learning method, it is also crucial that it can discover the best target features automatically without \textit{a priori} knowledge. 

\subsection{Overall Algorithm}
The overall GP-FRM algorithm is shown in \cref{gpfrmAlg}. A core component to the algorithm is the use of \textit{speciation}: the population is split into a number of \textit{species}, each of which share a common target feature. This niching approach serves two main purposes: it encourages multiple diverse FRs to be produced in a single GP run (rather than only a single best individual), while also restricting crossover and mutation to occurring only between individuals that share the same target feature, improving learning efficacy. The target feature of a given GP individual is automatically determined based at each generation, based on the feature which gives the best fitness, i.e.\ the first feature according to \cref{closestFeature}. This allows GP individuals to change more readily over time, moving between species or discovering an entirely new species niche.
\vspace{-.5em}

\begin{align}
	\label{closestFeature}
	ClosestFeatures(x|F) = \argsort_{f \in F}  |r_{x,f}| && (\textit{Decreasing sort})
\end{align}
\vspace{-.5em}


\begin{algorithm}[t]
	\caption{Overall GP-FRM Algorithm}
	\label{gpfrmAlg}
	\begin{algorithmic}[1]
		\renewcommand{\algorithmicrequire}{\textbf{Input:}}
		\renewcommand{\algorithmicensure}{\textbf{Output:}}
		\REQUIRE Dataset: $X$, maximum generations: $G$, num species $N_S$
		\ENSURE Set of $S$ Feature Relationships
		\STATE $F \leftarrow X^{T}$
		\STATE Randomly initialise population $P$
		\FOR {$i = 1$ to $G$}
		    \FOR {$j = 1$ to $|P|$}
		    \STATE $P_{j}CF \leftarrow ClosestFeatures(P_{j},F)$ using \cref{closestFeature}
		    \STATE $P_{j}\text{Fitness} \leftarrow Fitness(P_{j}, P_{j}CF_0)$ using \cref{fitness}
		    \ENDFOR
		    \STATE $Species \leftarrow Speciate(P,N_S)$ using \cref{speciationAlg} \label{sp1}
		    \STATE $P_{new} \leftarrow \{\}$ 
		    \FOR{$SP \in Species$}
		    	\STATE $Offspring \leftarrow Breed(SP)$
		    	\STATE $Offspring.append(SP.seed)$
		    	\STATE $P_{new}.append(Offspring)$
		    \ENDFOR \label{sp2}
		\ENDFOR
 		\FOR {$j = 1$ to $|P|$}
 			\STATE $P_{j}CF \leftarrow ClosestFeatures(P_{j},F)$ using \cref{closestFeature}
 			\STATE $P_{j}\text{Fitness} \leftarrow Fitness(P_{j}, P_{j}CF_0)$ using \cref{fitness}
 		\ENDFOR
	    \STATE $Species \leftarrow Speciate(P,N_S)$ using \cref{speciationAlg}
		\STATE $S \leftarrow \{\}$
  		\FOR{$SP \in Species$}
  			\STATE $S.append(SP.seed)$
  		\ENDFOR
  		\RETURN $S$
	\end{algorithmic} 
\end{algorithm}
\begin{algorithm}[t]
	\caption{Speciation Algorithm}
	
	\label{speciationAlg}
	\begin{algorithmic}[1]
		\renewcommand{\algorithmicrequire}{\textbf{Input:}}
		\renewcommand{\algorithmicensure}{\textbf{Output:}}
		\REQUIRE Population: $P$, num species $N_S$
		\ENSURE Set of $Species$ 
		\STATE $PSorted \leftarrow Sort(P)$
		\STATE $Species \leftarrow []$
		\FOR {$j = 1$ to $|PSorted|$}
			\STATE $S_{Index} \leftarrow PSorted_{j}CF_0$
			\IF{$S_{Index} \in Species$}
				\STATE $Species[S_{Index}].append(PSorted_{j})$
			\ELSIF{$|Species| < N_S$}
				\STATE$Species[S_{Index}].seed \leftarrow PSorted_{j}$
				\STATE $Species[S_{Index}].append(PSorted_{j})$
			\ELSE
				\STATE $k \leftarrow 1$
				\WHILE{$S_{Index} \notin Species$}
					\STATE $k \leftarrow k+1$
					\STATE $S_{Index} \leftarrow  PSorted_{j}CF_k $
				\ENDWHILE
				\STATE $Species[S_{Index}].append(PSorted_{j})$
			\ENDIF
		\ENDFOR
		\RETURN $Species$
	\end{algorithmic} 

\end{algorithm}

The core of \cref{gpfrmAlg} is similar to a standard evolutionary search, with the main difference being the use of speciation for breeding and elitism (Lines \ref{sp1}--\ref{sp2}). The speciation algorithm is shown in \cref{speciationAlg}. The number of species ($N_S$) is a parameter of the algorithm, and is used to constrain the number of niches in the search space: having too many species would give many poorer-quality FRs and prohibit niche-level exploitation by a group of individuals. The population is sorted by fitness (best to worst) into a Closest Features ($CF$) list and then each individual is considered in turn: 

\begin{enumerate}
	\item if the individual's closest feature ($CF_0$) has already been selected as a species, it is added to that species;
	\item otherwise, if the number of species ($N_S$) has not been reached, a new species is created with the individual's closest feature ($CF_0$) as the \textit{seed};
	\item otherwise, the individual's list of closest features ($P_{j}CF$) is searched to find the first seed (feature) which is in the species list, and then the individual is added to that species.
\end{enumerate}

In this way, the species are always selected from the fittest individuals, and the species seed represents the best individual in that species. A species' seed is always transferred to the next generation unmodified during the breeding process, as a form of elitism. When breeding a species, the number of offspring produced is $\frac{|P|}{N_S}$ to ensure each species has equal weighting.

\subsection{Fitness Function}
\label{ffSec}

A simple approach to assess the quality of a tree would be to measure the error between its output and its target feature, for example, the mean absolute error (MAE):
\begin{equation}
	MAE(x,f) = \frac{\sum_{i=1}^{|x|} |x_i - f_i|}{|x|}
\end{equation}
where $x$ is the $n$-dimensional output of a given tree, $f$ is the $n$-dimensional target feature, and $n$ is the number of instances.

The MAE is sensitive to scale: if $x$ was exactly 10 times the scale of $f$, it would give an error of $9$. This presents two problems: firstly, it means that the GP algorithm must learn constant factors within a FR, which traditional GP algorithms struggle with due to their use of random mutation\footnote{For example, mutating the $0.71$ node of $x = f_1 \times (f_0 + 0.71)$ using a traditional mutation would give a new value in $U[0,1]$. While local-search approaches can be used to optimise constants more cautiously, it is best if they can be avoided completely.}. Secondly, the scale of the learnt FRs is not actually important in many cases: a relationship between weight and height, for example, is meaningful whether weight is measured in grams, kilograms, or pounds. With these issues in mind, we instead employ Pearson's correlation, $r_{x,f}$, as our cost measure, given its scale invariance:

\begin{equation}
	\label{pearsons}
	\text{Cost} = r_{x,f} = \frac{
		\sum_{i=1}^{n}(x_i - \overline{x})(f_i - \overline{f})}
	{\sqrt{\sum_{i=1}^{n}(x_i-\overline{x})^2 } \sqrt{\sum_{i=1}^{n}(f_i-\overline{f})^2}
	}
\end{equation}

Pearson's correlation has a value between $+1$ and $-1$, where a value of $+1$ represents a completely positive linear relationship from $x$ to $f$, $0$ represents no correlation as all, and $-1$ represents a completely negative linear relationship. The magnitude of the correlation measures the degree of linearity in the relationship; the sign provides the directionality. We do not consider the directionality to be important in this work, as a negative feature relationship is equally as informative as a positive one. Therefore, we consider the absolute value of $r_{x,f}$ which is in the range $[0,1]$, where $1$ is optimal. Pearson's correlation has seen previous use in GP to encourage diversity and approximate fitness \cite{tomassini2005study,haeri2017statistical}.

If an evolved FR is to be realistically useful in understanding data, it must be sufficiently small and simple for a human to easily interpret. To achieve this, we introduce a penalty term into the fitness function, with an $\alpha$ parameter that controls the trade-off between high correlation (high cost) and small tree size. In practice, $\alpha$ is generally small, so this can be seen as a relaxed version of lexicographic parsimony pressure \cite{luke2002lexicographic}. The proposed fitness function --- which should be \textit{minimised} --- is shown in \cref{fitness}, for an individual $x$ with target feature $f$.

\begin{equation}
	\label{fitness}
	Fitness(x|f) = \begin{cases}
		1+|r_{x,f}|+\alpha \times size(x_{\text{Tree}}),& \text{if } f \in x_{\text{Tree}}\\
		1-|r_{x,f}| +\alpha \times size(x_{\text{Tree}}),              & \text{otherwise}
	\end{cases}
\end{equation}

\Cref{fitness} consists of two cases: one where the target feature $f$ is used in the tree $x_\text{Tree}$ and one where it is not. This is to penalise the evolution of na\"ive or self-referential trees such as $f_1 = f_1$ or $f_2 = f_1 \times \frac{f_2}{f_1}$. In the case where a target feature \textbf{is} used in $x_\text{Tree}$, the fitness is penalised by the size of the linear correlation (i.e.\ in the range $[1,2]$, disregarding $\alpha$). When it is not, the fitness will be in the range $[0,1]$. In this way, the fitness of an individual not using the target feature will always be better (lower) than another that does.

\subsection{Preventing the Discovery of Na\"{i}ve Relationships}
Often in many real-world datasets, features will be highly linearly correlated with each other: either due to redundancy in the feature set, or due to other natural linearity. For example, weight measured as a feature in $kg$ will be perfectly correlated with weight measured in $lb$. While it is not incorrect for GP to discover such relationships, they are not very useful, as they can be found in $O(n^2)$ time for $n$ features, by calculating the pairwise Pearon's correlation matrix. 

We prevent GP from evolving such relationships by pre-computing a list of ``matching features'' for each feature. This list contains all the other features that are linearly correlated with the feature\footnote{Two features are defined to be linearly correlated if they have an absolute Pearson's correlation greater than $0.95$.}. This list is used in place of $f$ in the calculation of fitness (\cref{fitness}), such that the fitness will be penalised if \textit{any} matching features to the target feature appear in the GP tree. Note that this does not prevent any features from being used as a species seed.

\section{Experiment Design}
To evaluate the potential of GP-FRM, we tested it on a range of real-world classification datasets (from different domains), which were selected due to having clearly human-meaningful features. These are summarised in \cref{table:datasets}, ordered according to the number of features. Some minor data cleaning was done, including the removal of missing values by removing whole features or instances as appropriate.

\begin{table}[tb]
\vspace{-1em}
	\caption{Classification datasets used for experiments.}
	\label{table:datasets}
	\centering
	\begin{tabularx}{.75\linewidth}{Xrrrr}
		\toprule
		Dataset & Features & Instances & Classes & Source \\
		\midrule
		Wine & 13 & 178 & 3 & \cite{uci}\\
		WDBC & 30 & 569 & 2 & \cite{uci} \\
		Dermatology & 34 & 358 & 6 & \cite{uci}\\
		Steel Plates Fault & 33 & 1941 & 2 & \cite{uci}\\
		PC3 & 37 & 1563 & 2 & \cite{shirabad2005steel}\\
		Spambase & 57 & 4601 & 2 & \cite{uci}\\
			Arrhythmia & 278 & 420 & 12 & \cite{uci}\\
		MFEAT & 649 & 2000 & 10 & \cite{uci}\\
		\bottomrule
	\vspace{-1em}
	
	\end{tabularx}
\end{table}

GP-FRM was tested at three $\alpha$ values ($0.01, 0.001, 0.0001$) to evaluate the trade-off between correlation and tree size. On each dataset, 30 runs of GP-FRM were performed for each value of $\alpha$. The parameter settings used for GP-FRM are shown in \cref{table:parameterSettings}. A reasonably high population size and number of generations was used due to the cheap computational cost of the fitness function. In practice, GP-FRM tended to convergence by about $400-500$ generations. A small maximum tree depth of six was used to encourage interpretable trees, which also further reduced the computational cost. The number of species, $N_S$, was set to 10 for all experiments based on initial tests. In future, we hope to allow the number of species to be dynamically determined during the evolutionary process.

	\begin{table}[tb]
\vspace{-1em}

	\centering
	\caption{GP Parameter Settings.}
	\label{table:parameterSettings}
	\begin{tabularx}{0.75\linewidth}{ll X ll}
		
		\toprule
		Parameter& Setting && Parameter & Setting\\
		\cmidrule(r){1-2}  \cmidrule(l){4-5}
		Generations & 1000 && Population Size & 1000\\
		Mutation & 20\% && Crossover & 80\% \\
		Selection & Tournament && Max.\ Tree Depth & 6\\
		Elitism & 1-per-species && Pop. Initialisation & Half-and-half\\
		
		\bottomrule
	\end{tabularx}%
	\vspace{-1em}

\end{table}

\section{Results}
The mean fitness, cost, and number of nodes over $30$ runs for each dataset and value of $\alpha$ are shown in \cref{meanResults}\footnote{Note that $\text{Fitness}=\text{Cost}+ \alpha \times\text{Nodes}$, but we also list the fitness separately for completeness.}. 

\begin{table}[htb]
\vspace{-1em}

\centering
\caption{Mean results of GP-FRM across the datasets}
\label{meanResults}
\begin{tabular}{@{}l S[scientific-notation = false] S[group-digits = false] S[group-digits = false] S@{}}
\toprule
Dataset     & {Alpha}       & {Fitness} &  {Cost} &  {Nodes} \\
\midrule
\multirow{3}{*}{Wine} & 0.0001 &    0.153 &     0.149 &     48.5 \\
& 0.0010 &    0.195 &     0.172 &     22.7 \\
& 0.0100 &    0.298 &     0.226 &     7.15 \\
\midrule
\multirow{3}{*}{WDBC} & 0.0001 &   0.0203 &     0.018 &     23.4 \\
& 0.0010 &   0.0322 &     0.023 &     9.17 \\
& 0.0100 &   0.0827 &    0.0403 &     4.24 \\
\midrule
\multirow{3}{*}{Dermatology} & 0.0001 &   0.0362 &    0.0334 &     27.9 \\
& 0.0010 &   0.0541 &     0.043 &     11.1 \\
& 0.0100 &    0.111 &    0.0631 &     4.76 \\
\midrule
\multirow{3}{*}{Steel Plates Fault} & 0.0001 &  0.00712 &   0.00552 &       16 \\
& 0.0010 &   0.0166 &   0.00906 &     7.55 \\
& 0.0100 &   0.0573 &    0.0225 &     3.48 \\
\midrule
\multirow{3}{*}{PC3} & 0.0001 & 0.000565 &  0.000162 &     4.04 \\
& 0.0010 &  0.00395 &  0.000334 &     3.62 \\
& 0.0100 &   0.0339 &   0.00379 &     3.01 \\
\midrule
\multirow{3}{*}{Spambase} & 0.0001 &     0.14 &     0.136 &     35.5 \\
& 0.0010 &     0.16 &     0.144 &     16.1 \\
& 0.0100 &    0.237 &     0.165 &     7.23 \\
\midrule
\multirow{3}{*}{Arrhythmia} & 0.0001 & 0.000298 &  1.63e-07 &     2.98 \\
& 0.0010 &  0.00299 &     1e-08 &     2.99 \\
& 0.0100 &   0.0298 &  4.07e-07 &     2.98 \\
\midrule
\multirow{3}{*}{MFEAT} & 0.0001 &  0.00955 &   0.00807 &     14.8 \\
& 0.0010 &   0.0166 &     0.011 &     5.54 \\
& 0.0100 &    0.043 &     0.013 &        3 \\
\bottomrule
\end{tabular}
\vspace{-1em}

\end{table}

As $\alpha$ is increased, there is a clear increase in mean cost and decrease in the mean number of nodes on most of the datasets in \cref{meanResults}. In many cases, the proportional decrease in the number of nodes is much higher than the increases in cost: for example, on the Wine dataset, the number of nodes at $\alpha=0.001$ is less than half that of at $\alpha=0.0001$, but the cost only increases from $0.149$ to $0.172$. Similar patterns are seen for WDBC, Dermatology, Spambase, and MFEAT. When $\alpha$ is increased from $0.001$ to $0.01$, however, the increase in cost is often proportionally much higher, especially on WDBC and PC3. The Arrhythmia dataset appears to exhibit strange behaviour, as a result of it having a high number of features which are simple multiplicative combinations of other features. Such relationships could be ``filtered out'' as a pre-processing step, using a similar approach to that of removing highly correlated features.

Generally, as the number of features in the dataset increases, the mean cost decreases. This is not surprising: the more features available, the more likely it is to find a stronger relationship between them. From a similar perspective, higher-dimensional datasets often require the use of fewer nodes, as there are a greater number of simple FRs to be found. A multiobjective GP approach \cite{badran2010influence} would likely help to find a balance between tree and/or node count and number of FRs.

To find more complex and interesting FRs, a greater number of species should be used on high-dimensional datasets. The $\alpha$ parameter should be set based on initial tests of cost: on datasets such as PC3, Steel Plates Fault, and MFEAT, a high $\alpha$ encourages small FRs while still achieving a very good cost. On other datasets such as Wine and Spambase, a low $\alpha$ is needed to ensure that the FRs found are of sufficient quality.

\begin{table}[tb]
\centering
\caption{Relative Standard Deviation of GP-FRM across the datasets}
\label{rsdResults}
\begin{tabular}{@{}l S[scientific-notation = false] S S S @{}}
\toprule
Dataset     & {Alpha}       & {Fitness} &  {Cost} &  {Nodes} \\
 & & {(\%)} & {(\%)} & {(\%)}\\
\midrule
\multirow{3}{*}{Wine} & 0.0001 &     41.7 &      42.8 &     24.1 \\
& 0.0010 &       37 &        40 &     31.3 \\
& 0.0100 &     28.9 &      31.4 &     42.1 \\
\midrule
\multirow{3}{*}{WDBC} & 0.0001 &     42.9 &      47.9 &     33.7 \\
& 0.0010 &     33.4 &      41.2 &     32.7 \\
& 0.0100 &     22.4 &      39.1 &     29.9 \\
\midrule
\multirow{3}{*}{Dermatology} & 0.0001 &     40.3 &      43.3 &     37.6 \\
& 0.0010 &       33 &      36.8 &     42.5 \\
& 0.0100 &     28.5 &      38.9 &     30.7 \\
\midrule
\multirow{3}{*}{Steel Plates Fault} & 0.0001 &     94.9 &       113 &     55.2 \\
& 0.0010 &       58 &      87.2 &     48.2 \\
& 0.0100 &     35.4 &      75.6 &     24.6 \\
\midrule
\multirow{3}{*}{PC3} & 0.0001 &       52 &       161 &     34.6 \\
& 0.0010 &     24.2 &       182 &     25.6 \\
& 0.0100 &     18.2 &       159 &     5.41 \\
\midrule
\multirow{3}{*}{Spambase} & 0.0001 &     61.1 &      62.2 &     43.4 \\
& 0.0010 &     56.9 &      60.8 &     48.4 \\
& 0.0100 &     43.8 &        59 &     33.1 \\
\midrule
\multirow{3}{*}{Arrhythmia} & 0.0001 &     6.76 &  1.73e+03 &     6.69 \\
& 0.0010 &     5.46 &  1.73e+03 &     5.46 \\
& 0.0100 &     6.69 &  1.17e+03 &     6.69 \\
\midrule
\multirow{3}{*}{MFEAT} & 0.0001 &     41.7 &      46.8 &     54.4 \\
& 0.0010 &       28 &      37.2 &     42.9 \\
& 0.0100 &       13 &      43.1 &        0 \\
\bottomrule
\end{tabular}
\end{table}

The relative standard deviation (RSD: $100\% \times \frac{SD}{mean}$) of these results is presented in \cref{rsdResults} to show the variation across the $30$ runs. In general, the RSD is around 20--50\%, aside from a few cases where it is much higher due to the measured values being very small. This level of variance is not unusual for GP, but could be reduced further in future work through the use of a more constrained search space, or the introduction of domain knowledge to give a higher-fitness initial population.

\begin{figure}[tbhp]
	\begin{subfigure}[b]{.5\textwidth}
  		\hbox{\hspace{-4em}\includegraphics[width=1.2\textwidth]{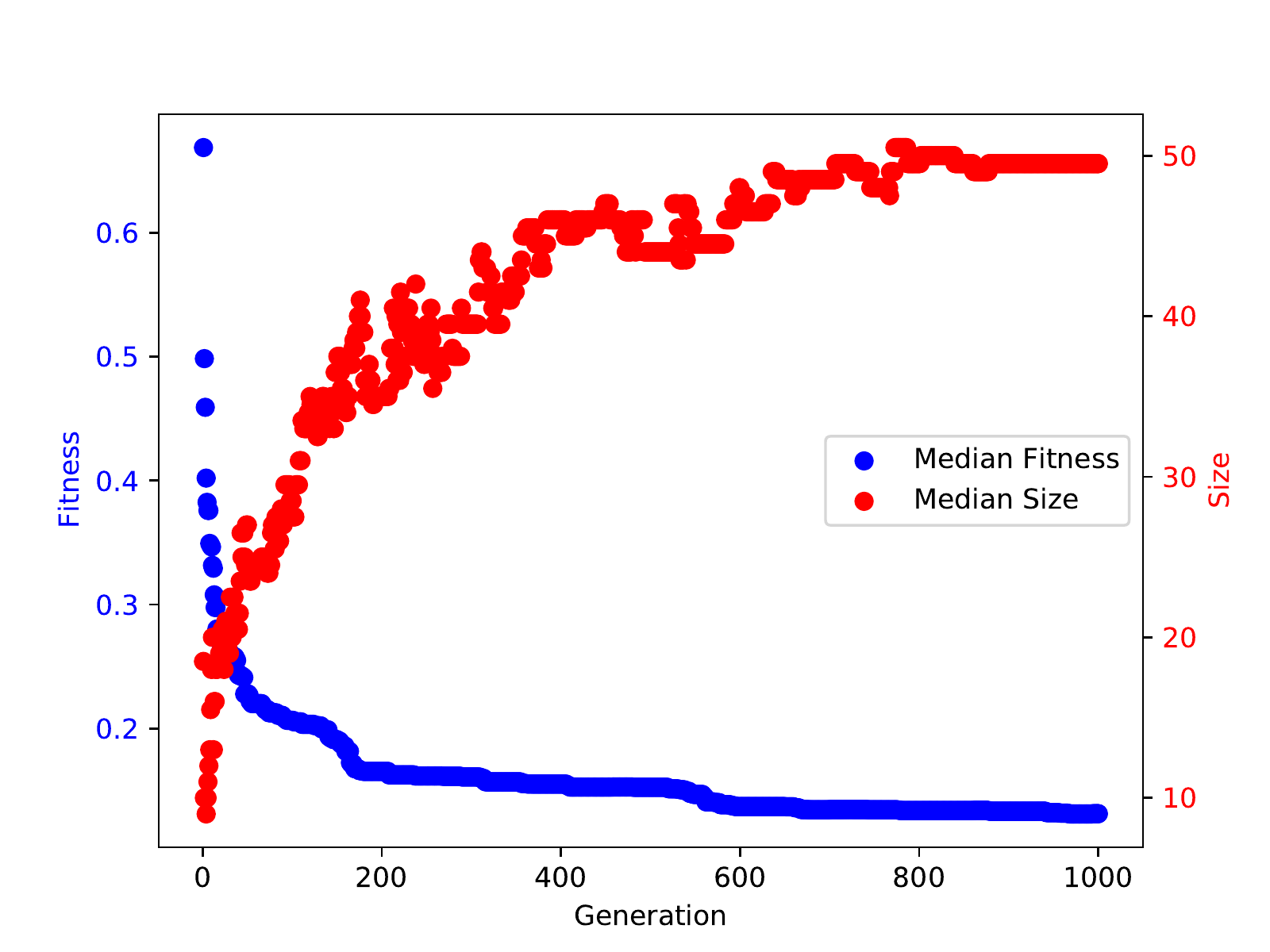}}
	\end{subfigure}%
	\hspace{-.5em}
	\begin{subfigure}[b]{.5\textwidth}
  		\includegraphics[width=1.2\textwidth]{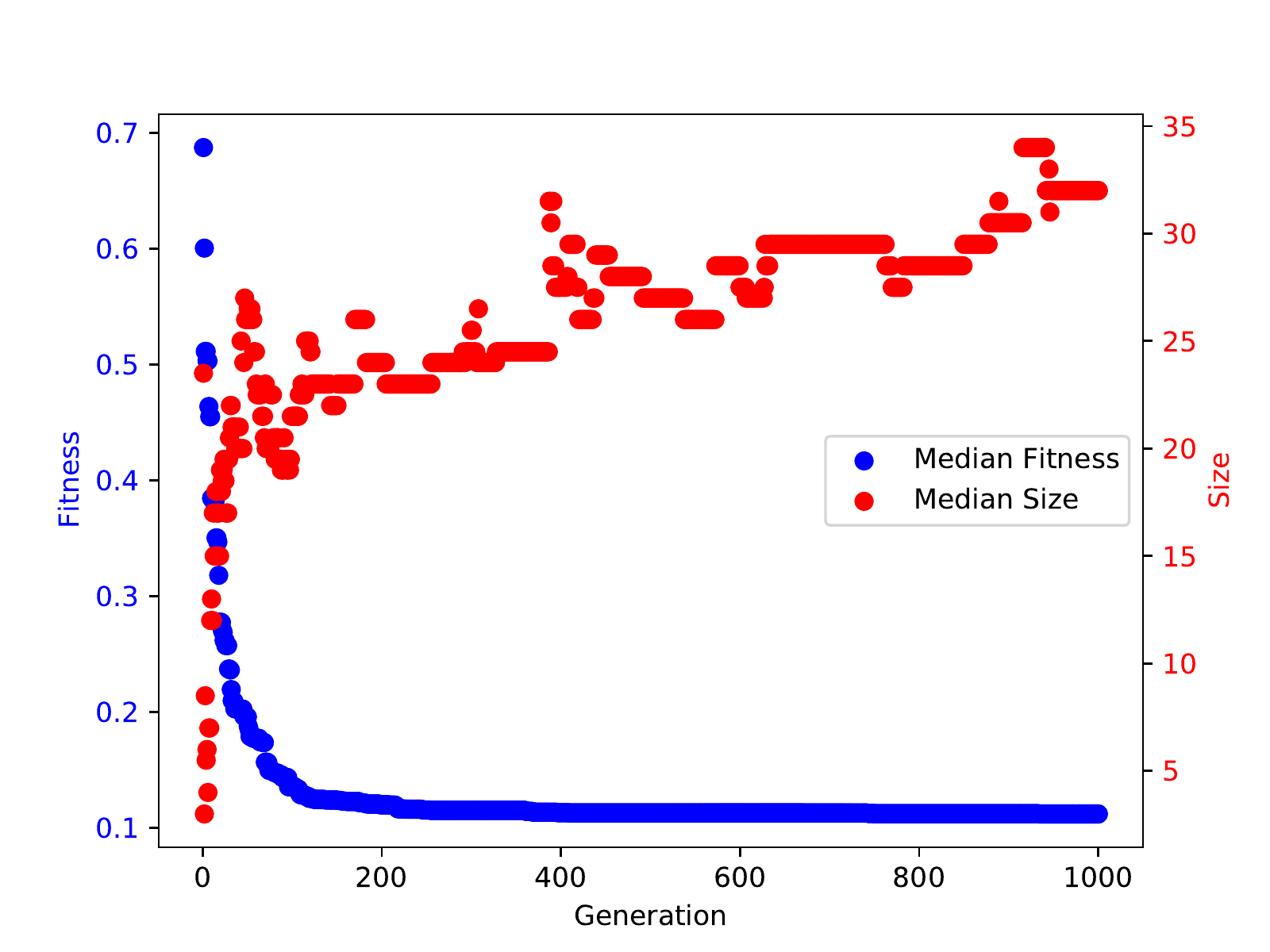}
	\end{subfigure}\\ %
	\begin{subfigure}[b]{.5\textwidth}
		\hbox{\hspace{-4em}\includegraphics[width=1.2\textwidth]{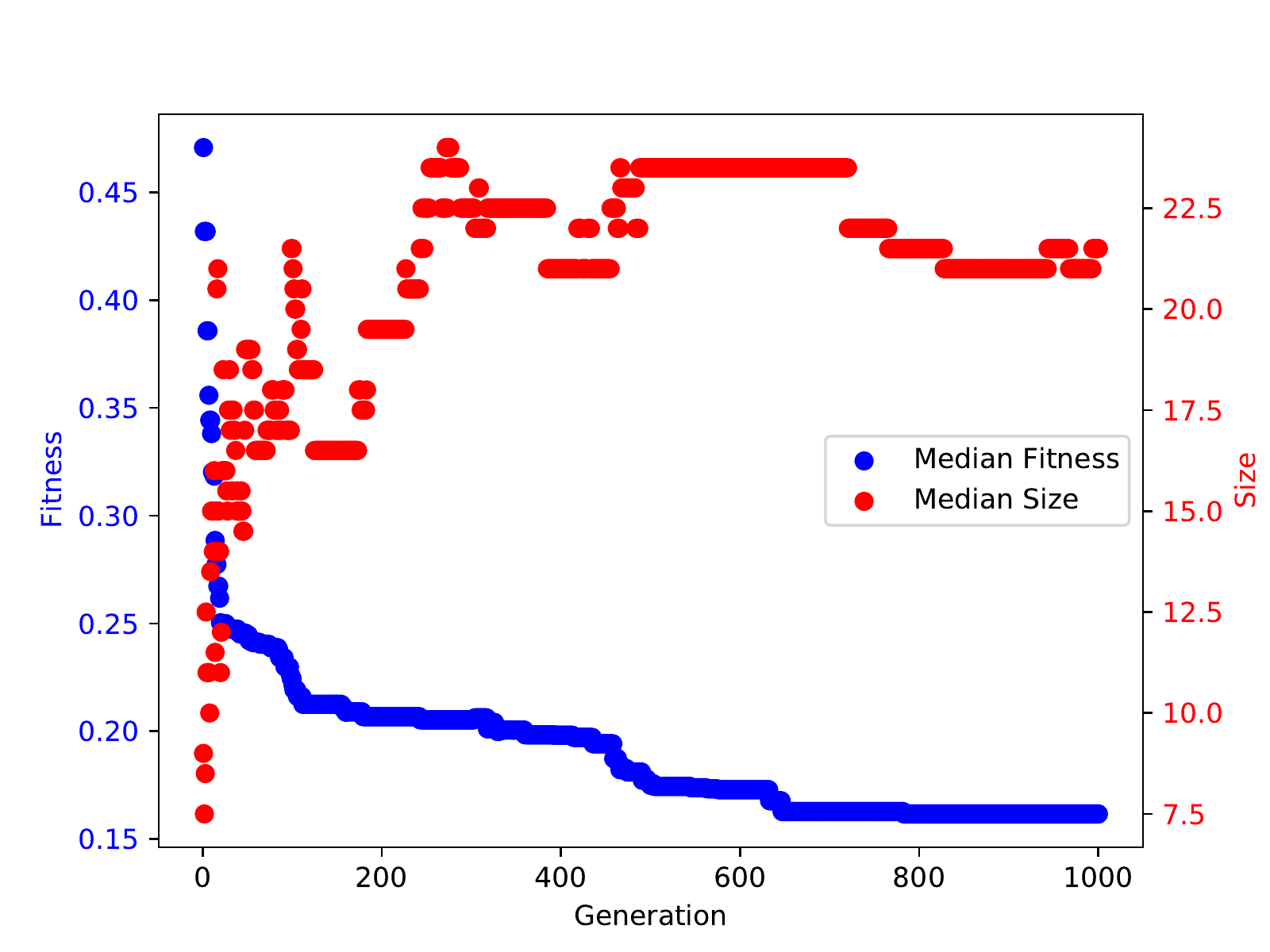}}
	\end{subfigure}%
	\hspace{-.5em}
	\begin{subfigure}[b]{.5\textwidth}
		\includegraphics[width=1.2\textwidth]{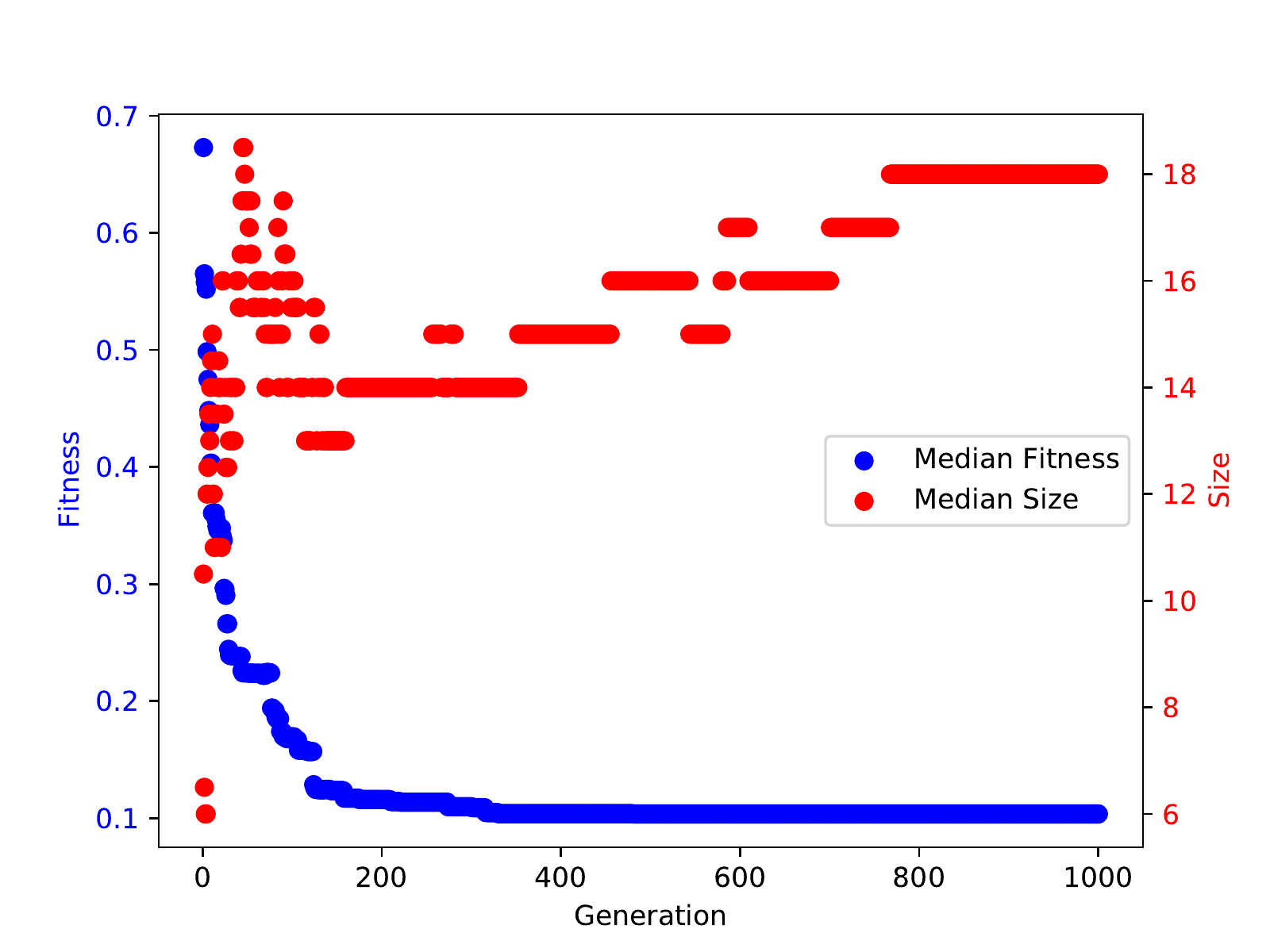}
	\end{subfigure}\\ %
	\begin{subfigure}[b]{.5\textwidth}
		\hbox{\hspace{-4em}\includegraphics[width=1.2\textwidth]{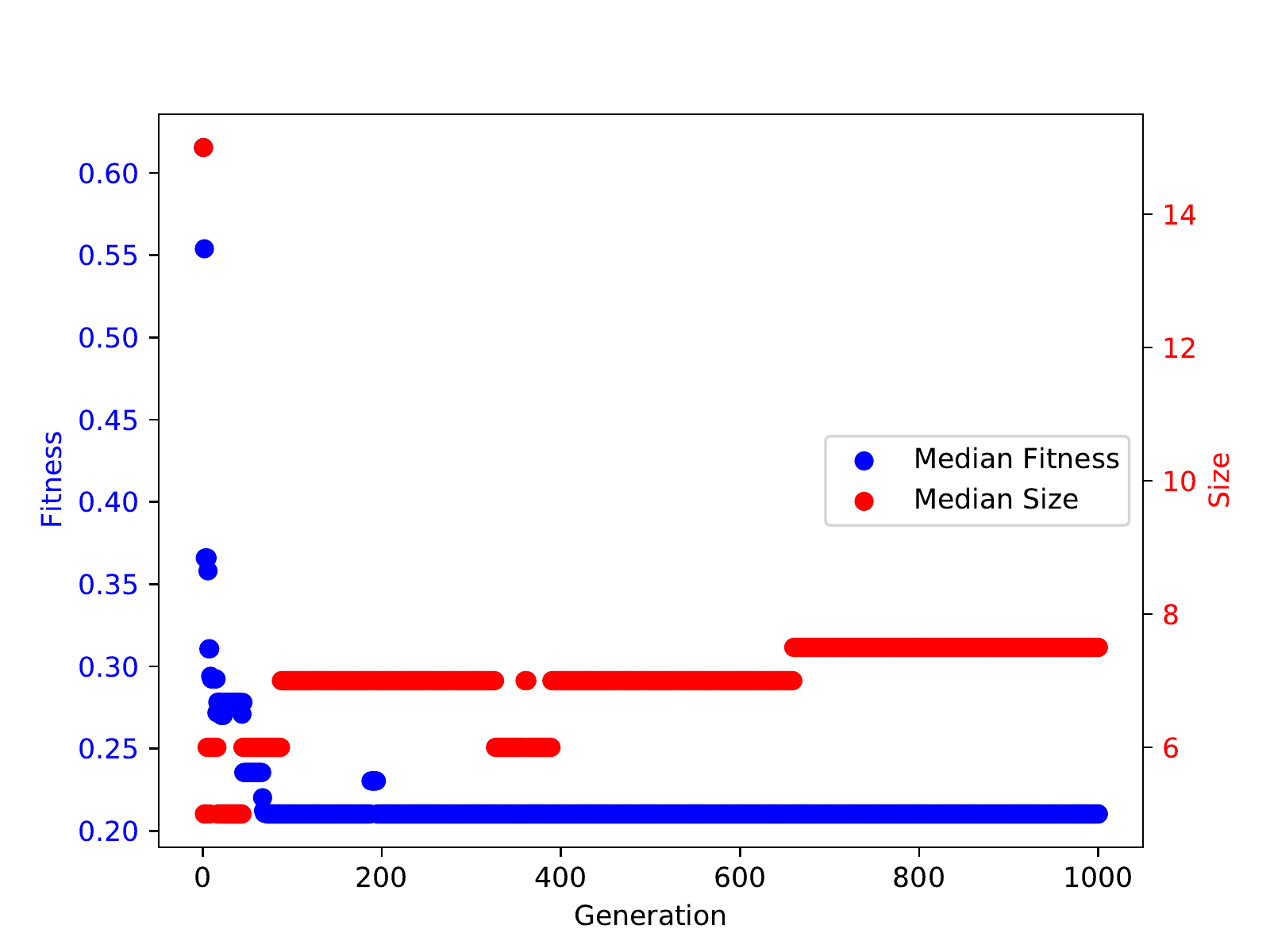}}
	\end{subfigure}%
	\hspace{-.5em}
	\begin{subfigure}[b]{.5\textwidth}
		\includegraphics[width=1.2\textwidth]{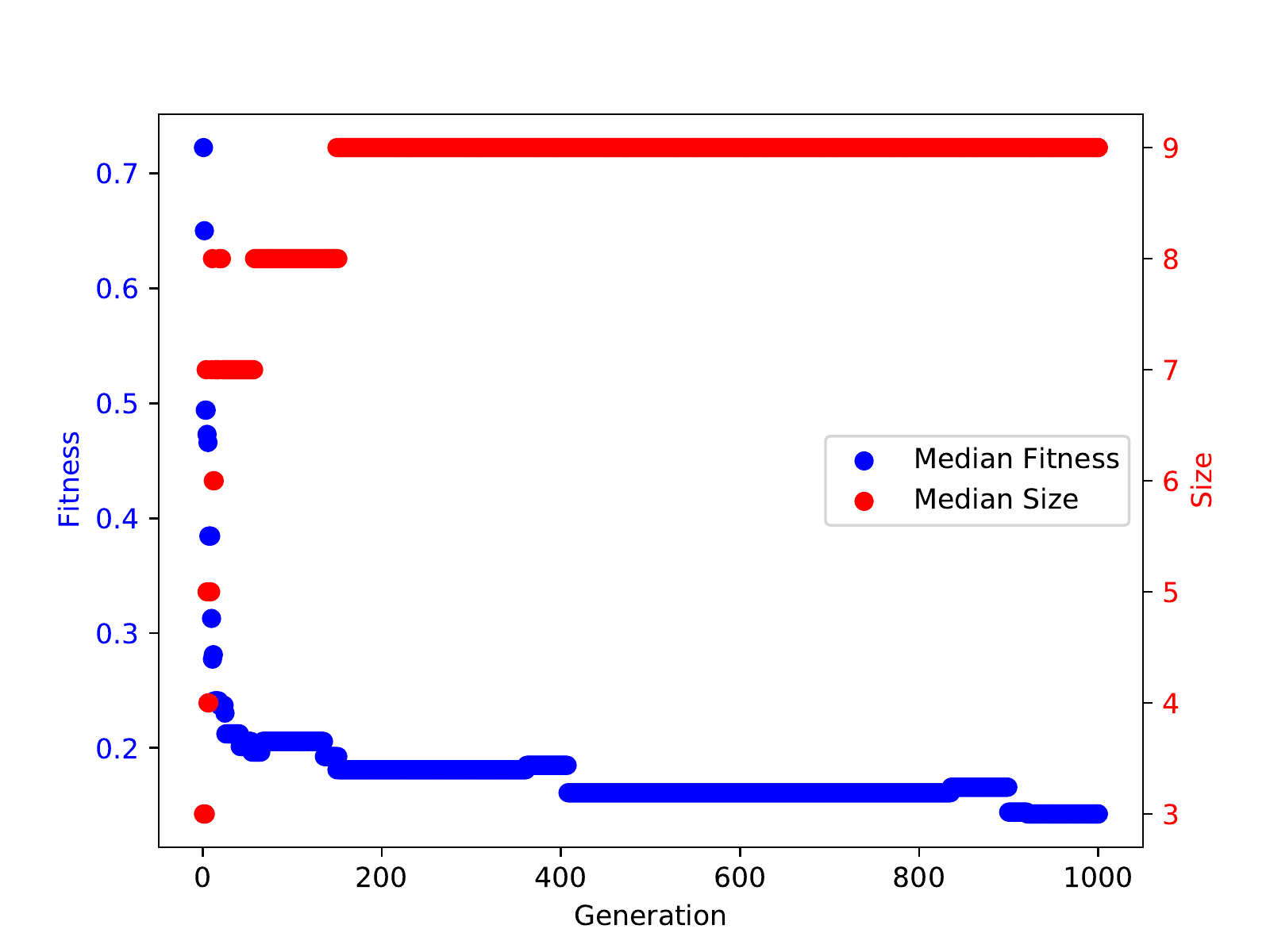}
	\end{subfigure}\\ %
	\caption{Convergence Analysis for the Wine (left) and Spambase (right) datasets. Top row is $\alpha=0.0001$, middle is $\alpha=0.001$, and bottom is $\alpha=0.01$. Median values of fitness and size are plotted to represent the expected average performance of a single GP run.}\label{convergenceFigs}
\end{figure}

\subsection{Convergence Curves}
The convergence curve for GP-FRM for each value of $\alpha$ is shown for two representative datasets (Wine and Spambase) in \cref{convergenceFigs}. Clearly, when $\alpha$ has a high value ($0.01$), convergence occurs much more quickly --- the high pressure to use few nodes in a tree greatly restricts the size of trees, reducing the number of good individuals in the search space. The convergence curve is also much less granular, due to the restrictions on tree size. At a lower $\alpha$, the size of individuals starts quite low, but then increases over the evolutionary process, before levelling off. This again reflects the difficulty of finding larger individuals which have sufficiently lower costs to out-perform simpler, but higher-cost individuals. Early in the evolutionary process, it is much "easier" to find small individuals that have a relatively good fitness than larger ones. In the future, it may be interesting to investigate dynamically adapting $\alpha$ throughout evolution, to better guide the search based on whether the cost or size of individuals is sufficiently low.

\section{Further Analysis}
To better understand the potential of GP-FRM for mining feature relationships which are useful for providing insight in data, we further analyse a selection of the evolved FRs in this section.

\subsection{Analysis of Selected Features}
\begin{figure}[p]
    \centering
    \vspace{-1.5em}
    \includegraphics[width=.95\textwidth]{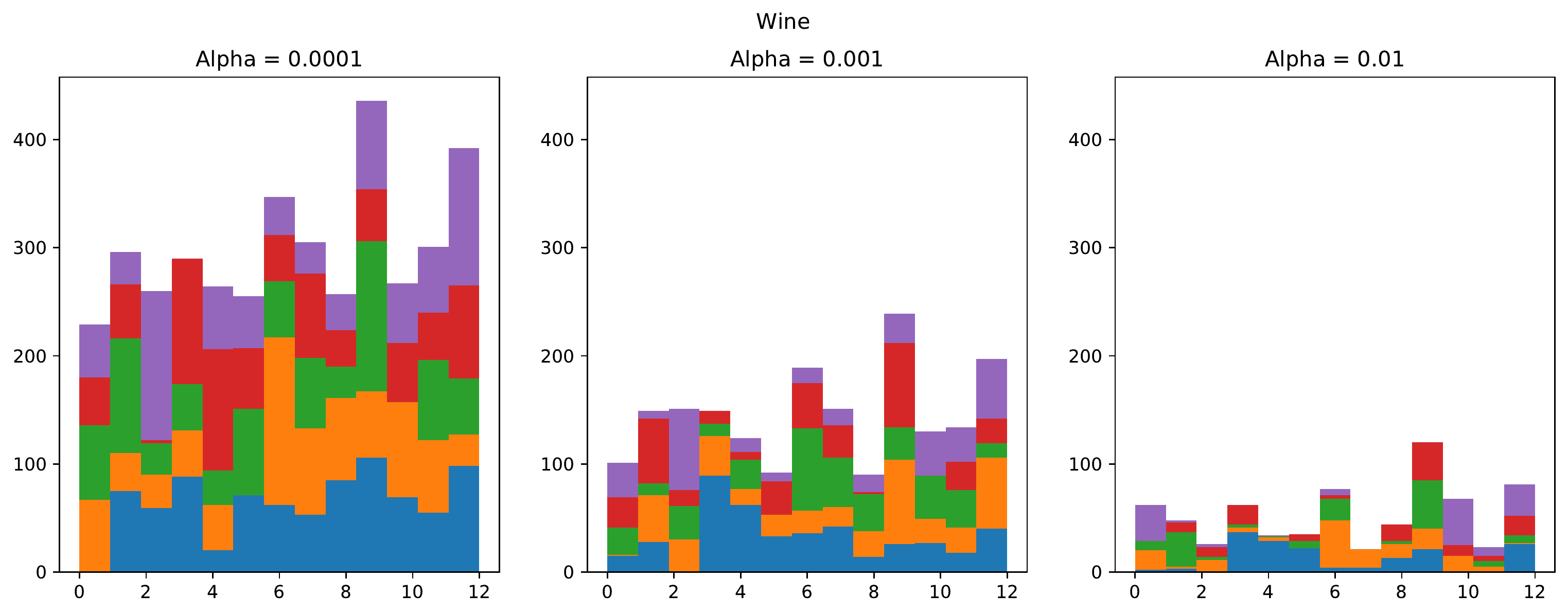}
    \includegraphics[width=.95\textwidth]{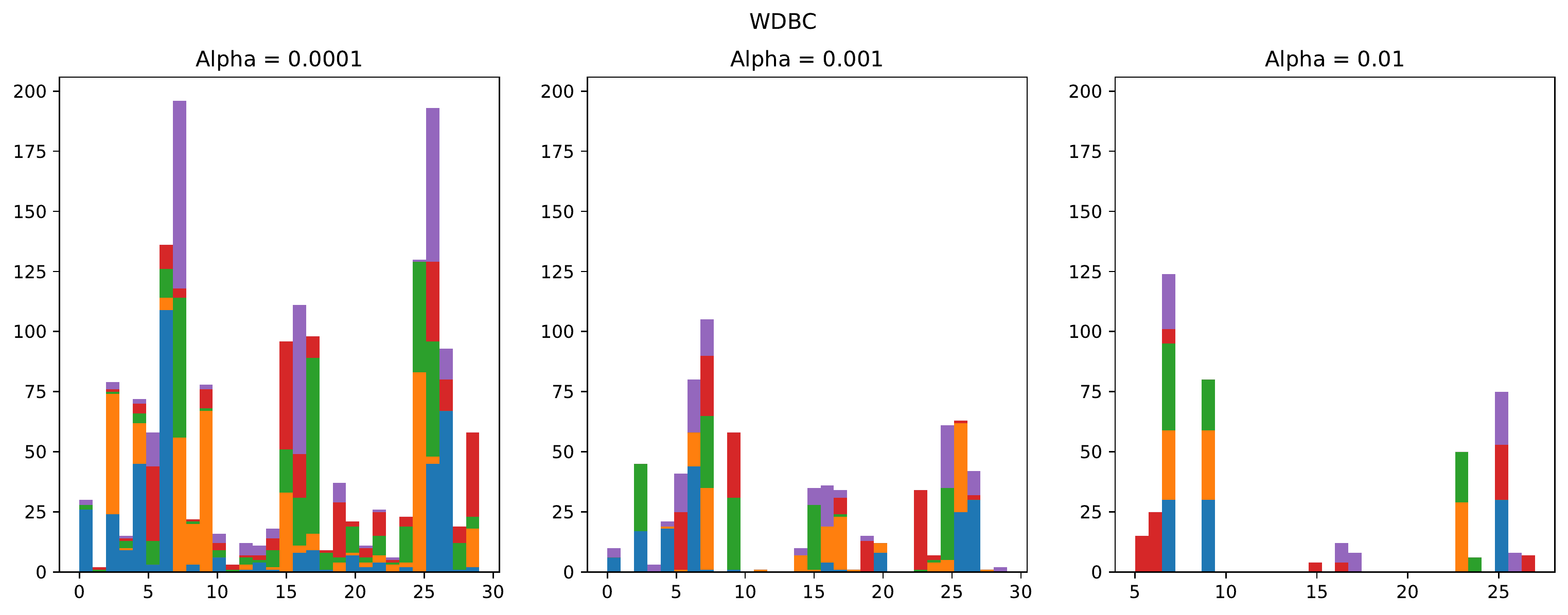}
    \includegraphics[width=.95\textwidth]{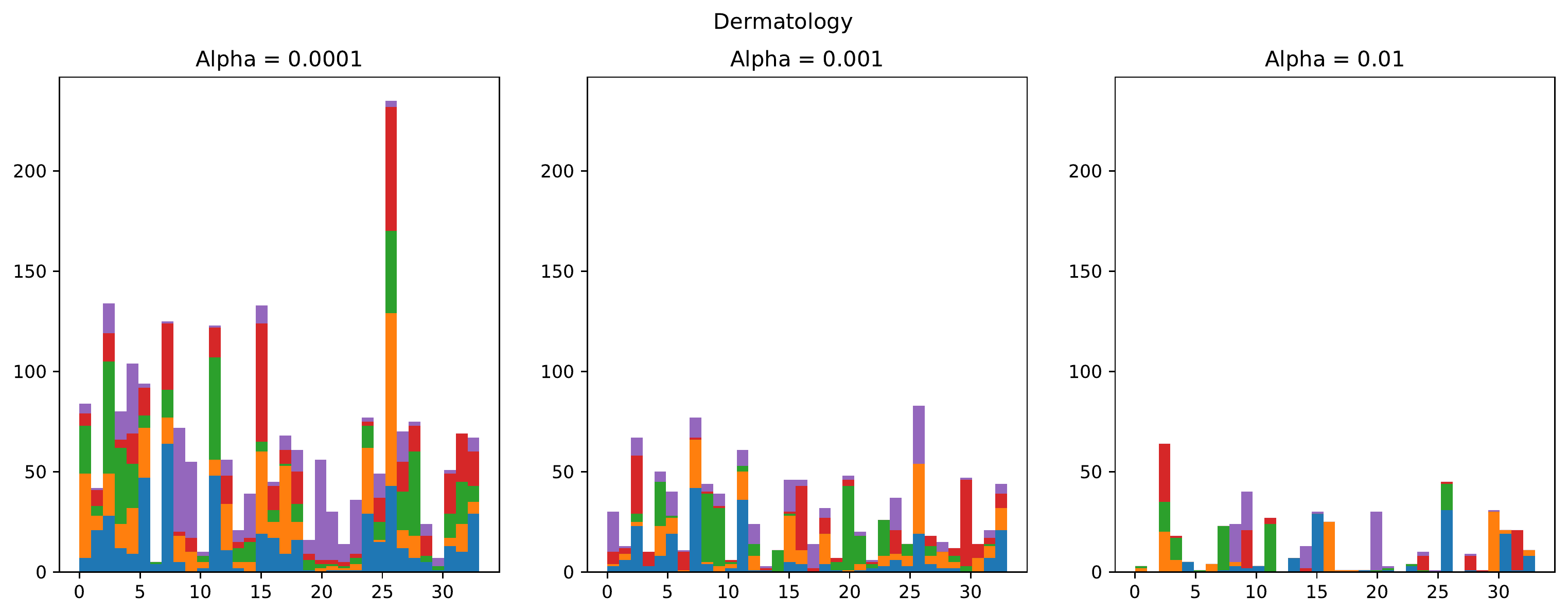}
    \includegraphics[width=.95\textwidth]{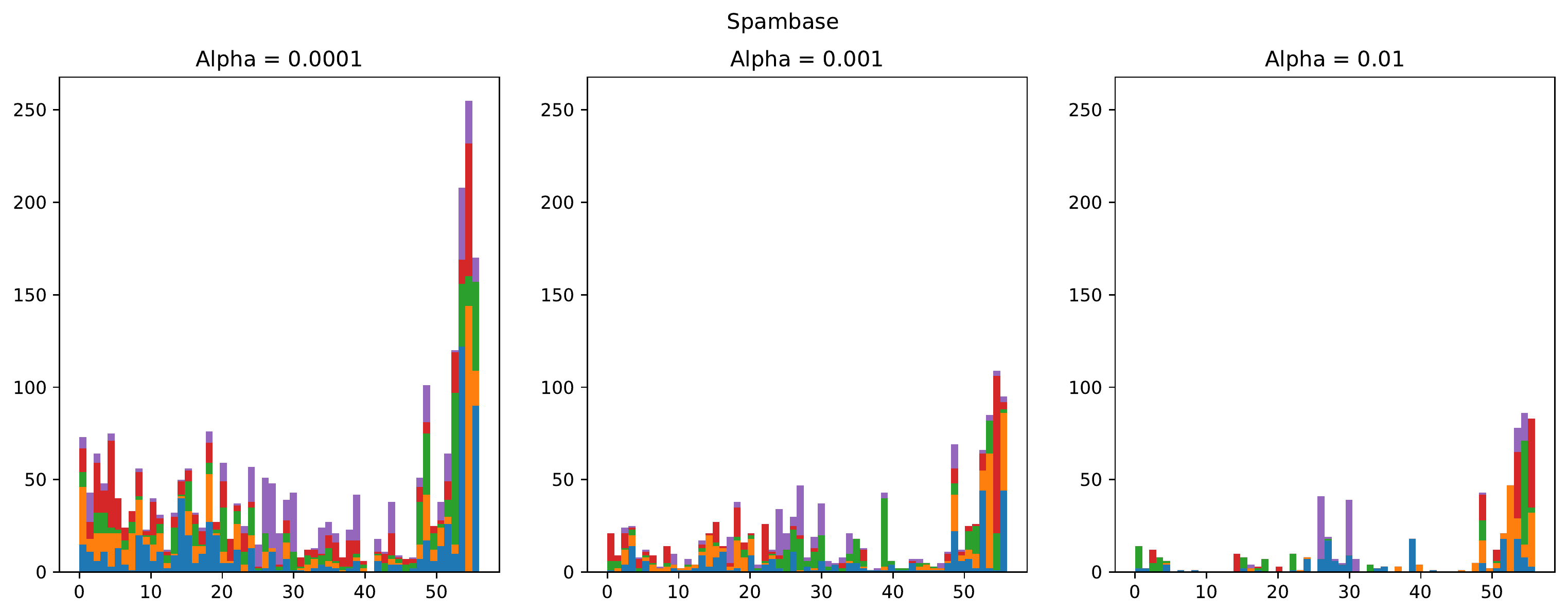}
    \caption{Histogram of the features used to produce feature relationships (FRs) for the five most common target features on four datasets. Each colour represents one target feature. $\alpha$ varies from left-to-right, increasing the penalty for using more nodes in a tree. The x-axis represents each feature indexed in the order it appeared in the dataset.}
    \label{fig:selectedFeatures}
\end{figure}

Of the eight tested datasets, the four which showed the biggest decrease in tree size from $\alpha=0.0001$ to $\alpha=0.01$ were Wine, WDBC, Dermatology, and Spambase. A decrease in tree size will generally give a decrease in the number of feature terminals, and hence decrease the occurrence of each feature in an evolved FR. This pattern can be seen in \cref{fig:selectedFeatures}, which plots the histogram of the features used to produce FRs for the five most common\footnote{Only the top five FRs are considered to make the plots easier to analyse.} target features in each of the datasets. As $\alpha$ is increased (from left to right), the number of features (the area under the histogram) decreases significantly. On three of the datasets, there are features which are never selected to be in a GP tree when $\alpha=0.01$. This shows that the parsimony pressure is not only encouraging GP to evolve smaller trees, but also simpler trees which use fewer distinct features. 

Across the four datasets, the five target features utilise clearly distinct groups of features. For example, at $\alpha=0.0001$ on Spambase, the purple target feature mostly uses features with indices between 25 and 30. There is also a clear spike on Spambase with features above index 50 being particularly popular for the blue and orange target features. On all datasets, we can see an increase in this niching-style behaviour as $\alpha$ is increased. For example, on the Wine dataset, at $\alpha=0.0001$, most features are commonly used across all the target features; at $\alpha=0.01$, only a few different features are commonly used for each target feature. This pattern reinforces the benefit of a speciation approach (particularly at a high $\alpha$) in encouraging multiple distinct FRs to be learned simultaneously. Using a more complex parsimony pressure that considers the number of unique features in a tree is likely to further improve the performance of speciation.

\begin{figure}[tbhp]
	\begin{subfigure}[b]{.49\textwidth}
		\includegraphics[width=\textwidth]{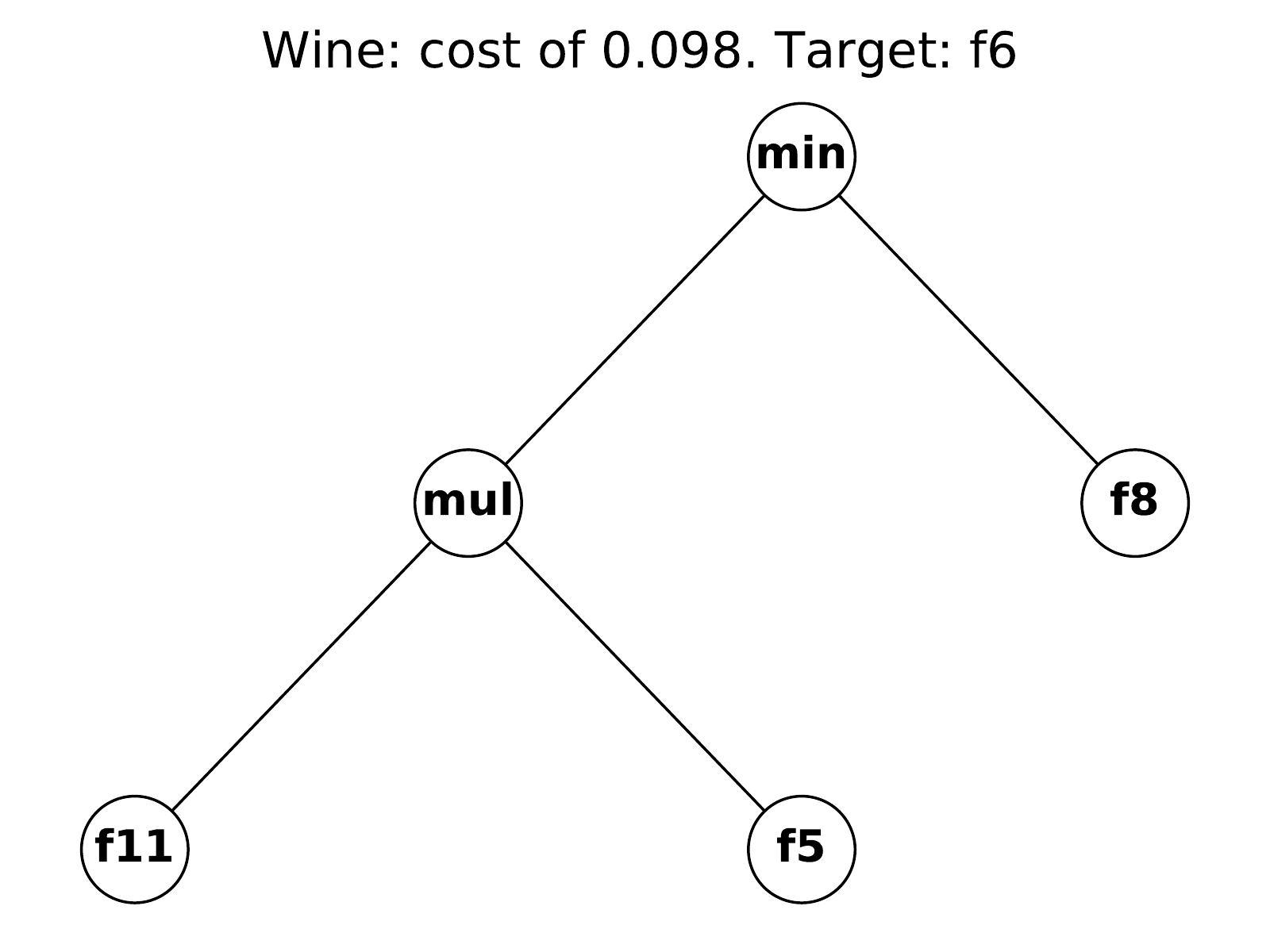}
	\end{subfigure} \hfill
	\begin{subfigure}[b]{.49\textwidth}
		\includegraphics[width=\textwidth]{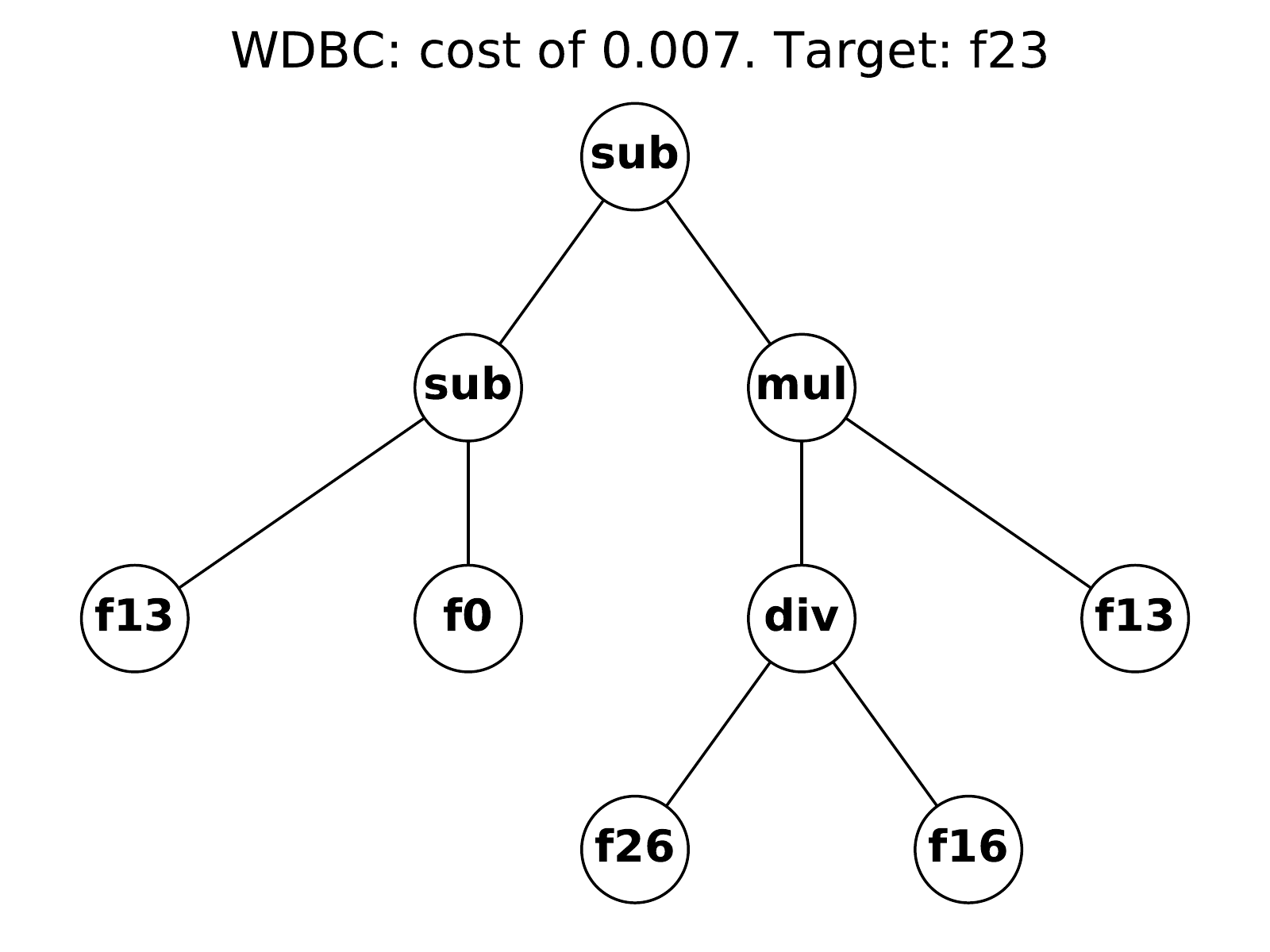}
	\end{subfigure}\\
	\begin{subfigure}[b]{.49\textwidth}
		\includegraphics[width=\textwidth]{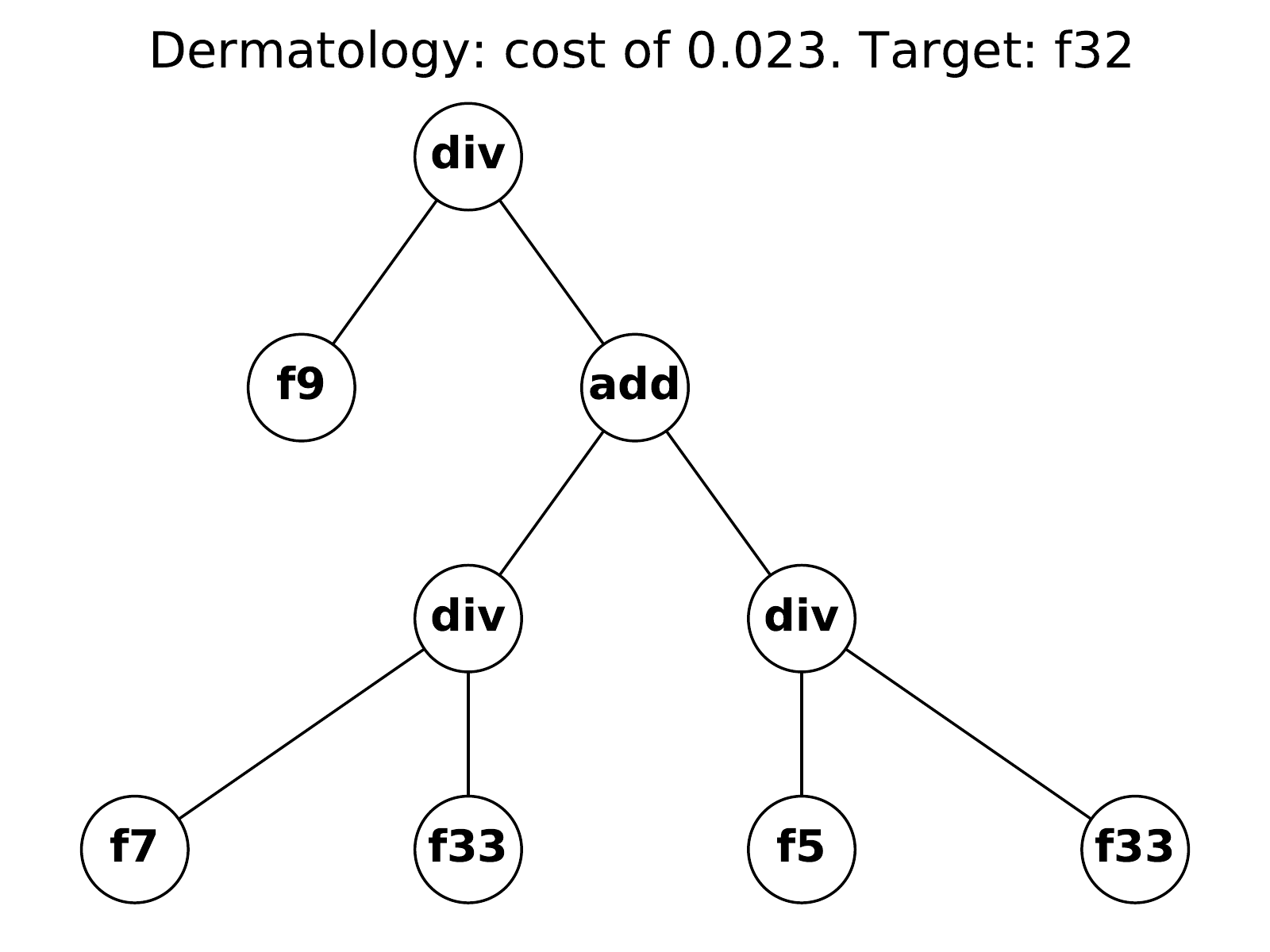}
	\end{subfigure} \hfill
	\begin{subfigure}[b]{.49\textwidth}
		\includegraphics[width=\textwidth]{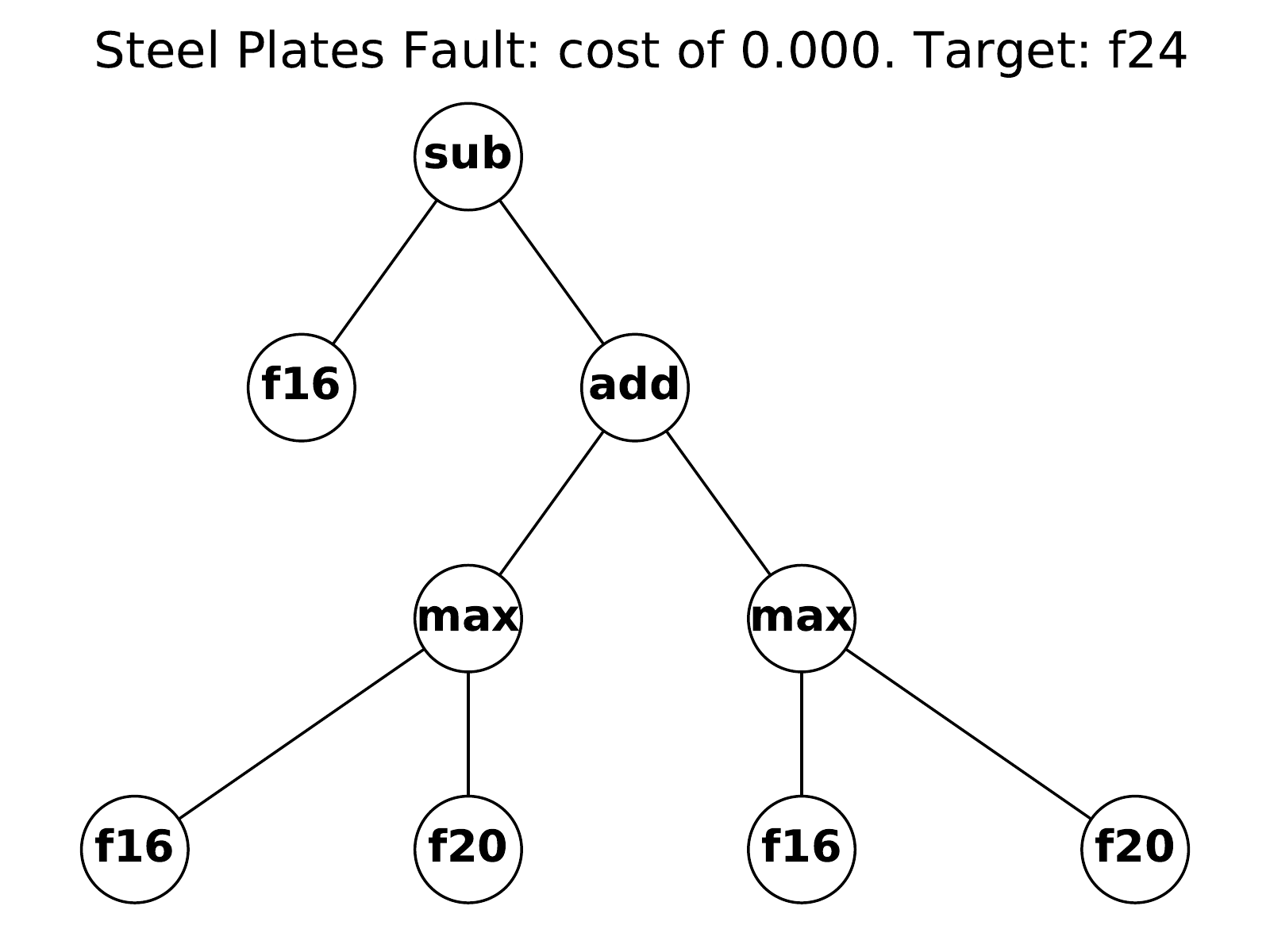}
	\end{subfigure}\\
	\begin{subfigure}[b]{.49\textwidth}
		\includegraphics[width=\textwidth]{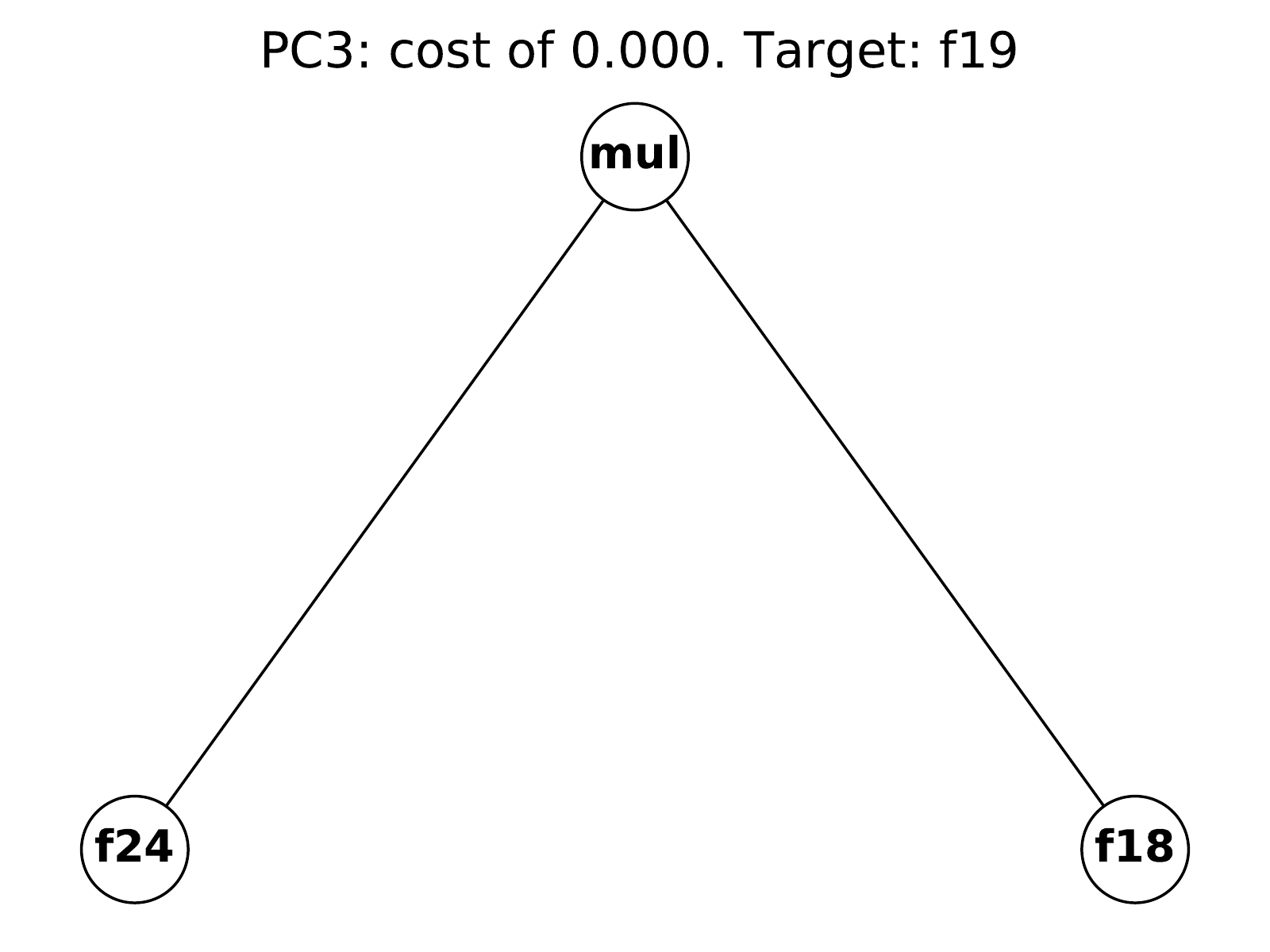}
	\end{subfigure} \hfill
	\begin{subfigure}[b]{.49\textwidth}
		\includegraphics[width=\textwidth]{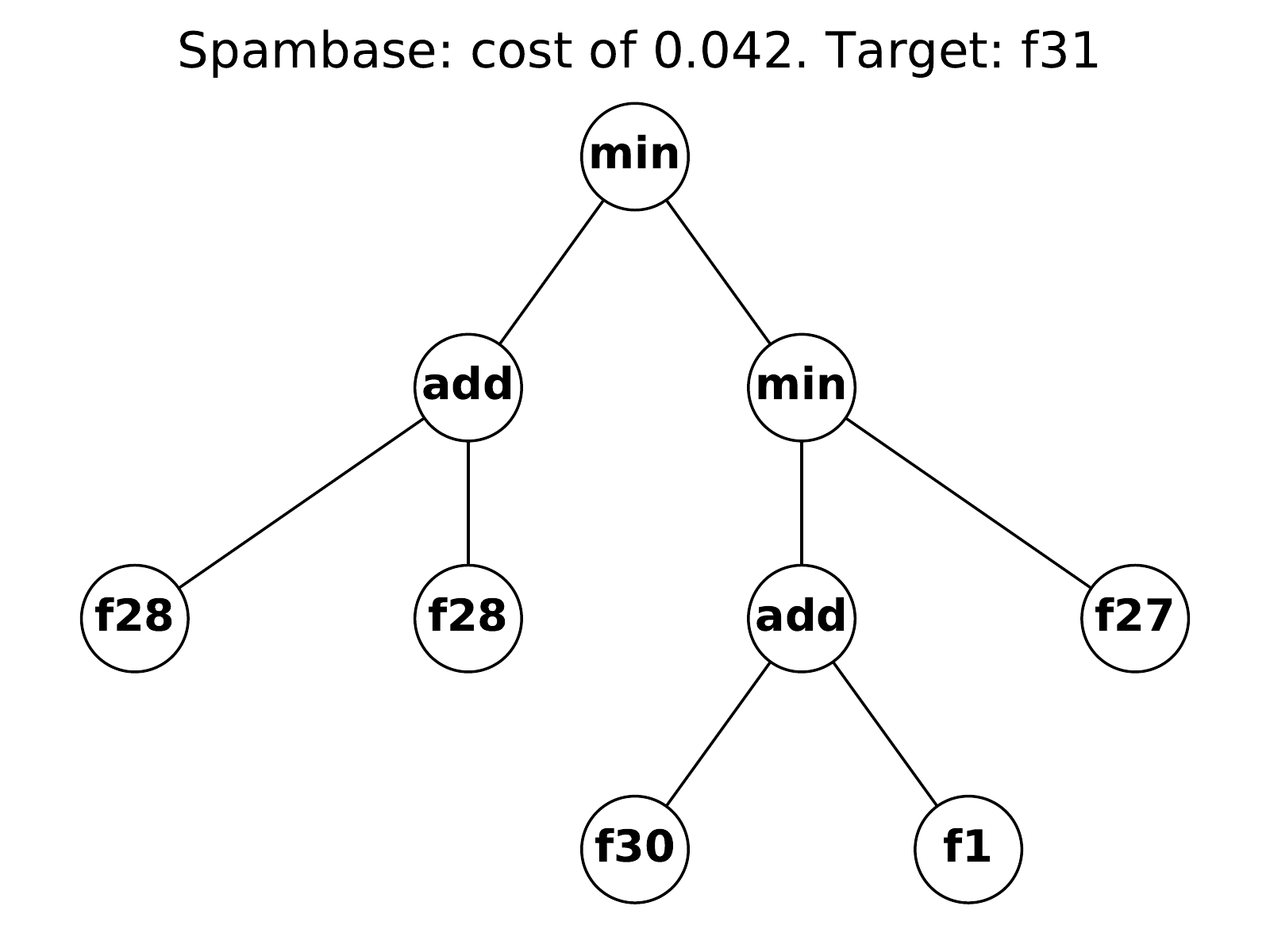}
	\end{subfigure}\\
	\begin{subfigure}[b]{.49\textwidth}
		\includegraphics[width=\textwidth]{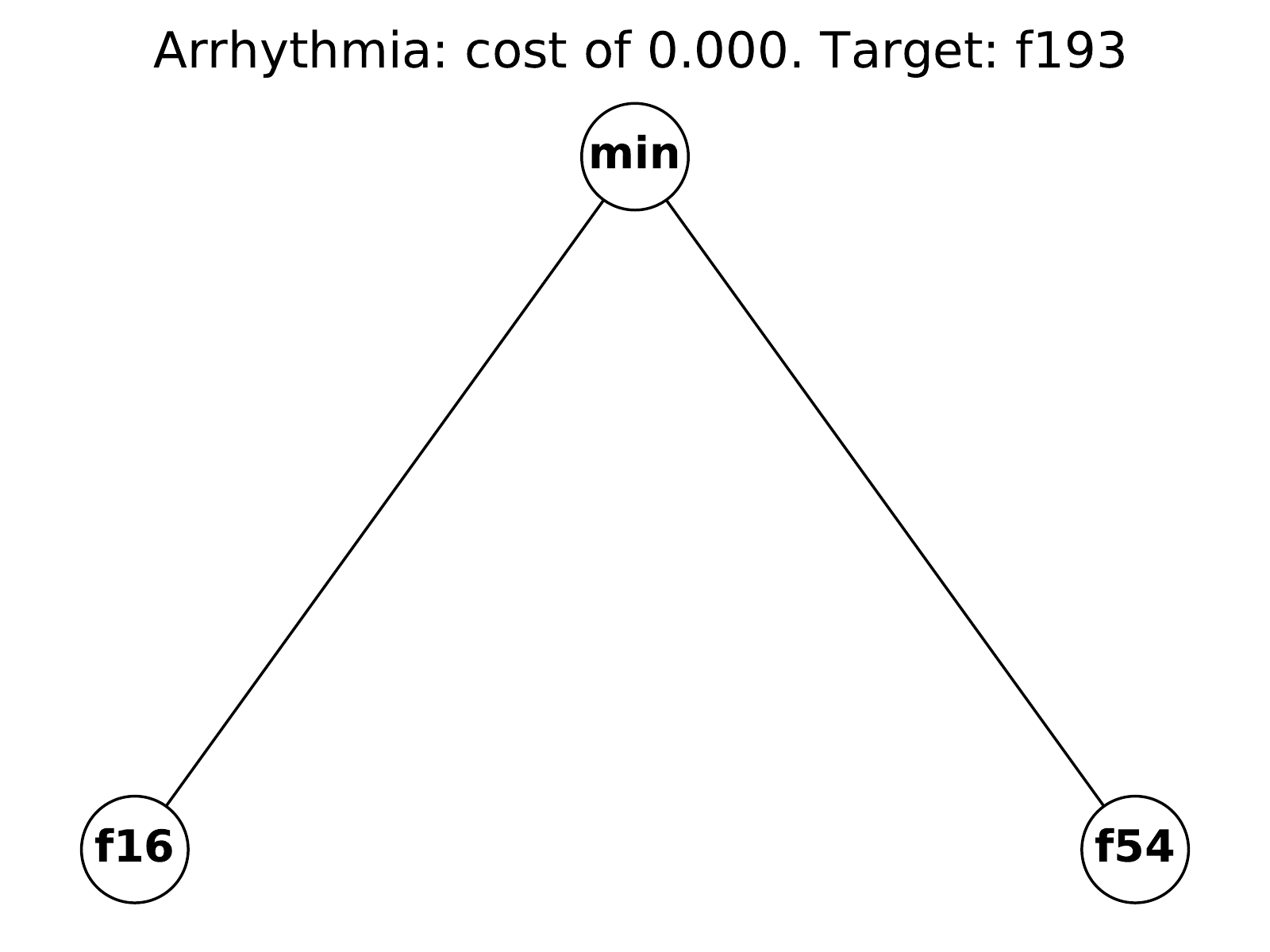}
	\end{subfigure} \hfill
	\begin{subfigure}[b]{.49\textwidth}
		\includegraphics[width=\textwidth]{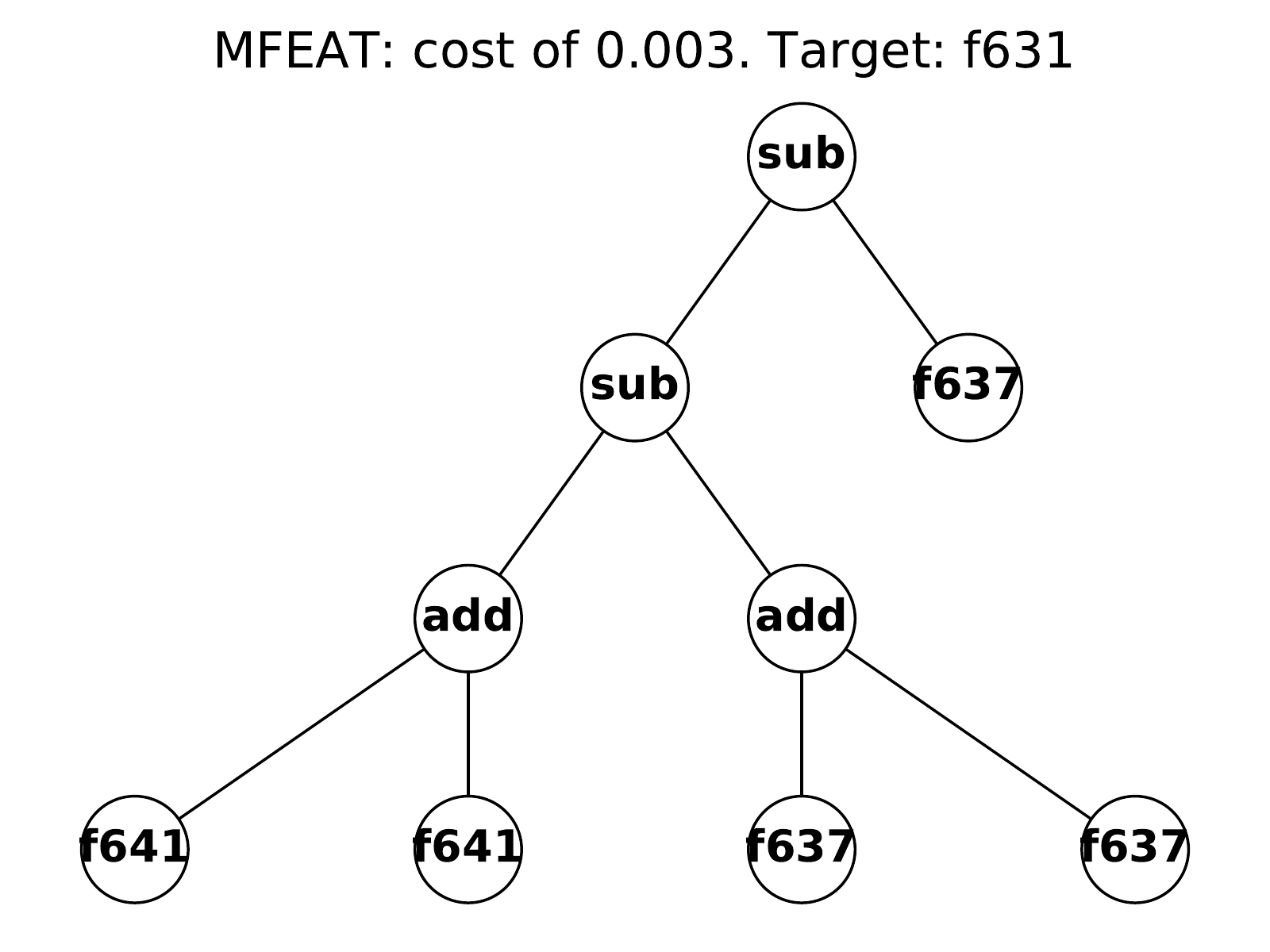}
	\end{subfigure}%
	
	\caption{Best evolved trees on each dataset that use fewer than five unique features and ten nodes in total.}\label{bestTrees}
\end{figure}

\subsection{Analysis of Evolved Relationships}
To further understand the usefulness of the evolved FRs, we selected the tree with the lowest cost on each dataset that used fewer than five unique features and no more than ten nodes overall. While other trees had slightly lower cost, their greater complexity makes them less useful for simple analysis. The eight trees for the eight datasets are shown in \cref{bestTrees}. We analyse a sample of these trees further to evaluate their meaning in the context of the features of the dataset.

The tree shown for the Wine dataset has the highest cost across the datasets, but a cost of $0.098$ still gives a Pearson's correlation of greater than $0.9$, indicating a very strong correlation \cite{cohen2013statistical}. $f5$, $f8$, and $f11$ correspond to ``total phenols'', ``proanthocyanins'' and ``hue'' respectively, with the target feature being flavanoids. This FR therefore shows that flavanoids have a high linear correlation with the greater of the amount of proanthocyanins and the product of total phenols and hue. This information could be useful to a food chemist in understanding how to control the amount of flavanoids in wine.

The target feature for WDBC is ``worst area'': the largest cell nucleus area in the breast tissue sample. The GP tree is equivalent to the formula: $\text{worst area} = \text{se area} - \text{mean radius} - (\frac{\text{worst concavity}}{\text{se concavity}}) \times \text{se area}$, where se is the standard error. 

The tree evolved on the Dermatology dataset uncovers a relationship between the presence of a band-like pattern on the skin, and other skin attributes, including a clear relationship with the age of the patient: \\$\text{band-like infiltrate} = \text{scalp involvement} \div(\frac{\text{oral mucosal involvement}}{\text{age}} + \frac{\text{polygonal papules}}{\text{age}})$. This could be very useful to dermatologists in understanding how the likelihood of different symptoms varies as a patient gets older.

On both the PC3 and Arrhythmia datasets, GP-FRM found very simple trees. For PC3, the discovered rule is $\text{Halstead Effort} = \text{Halstead Volume} \times \text{Halstead Difficulty}$.  The PC3 dataset measures various aspects of code quality of NASA software for orbiting satellites. The Halstead effort measures the ``mental effort required to develop or maintain a program'', and indeed is defined in the original paper in this formulation \cite{halstead1977elements}. The fact that GP-FRM discovered this (already known) relationship highlights its ability to find rules that ARM algorithms would not.

Finally, on the Spambase dataset, a high correlation is found between the number of times that the token ``857'' occurs in an email and a number of other tokens such as ``650'', ``telnet'', ``lab'', and ``address''. A security researcher analysing this dataset may be able to use this information to better understand common patterns in spam, in order to block it more accurately.

\section{Conclusion}
This paper proposed the first approach to automatically discovering feature relationships (FRs): symbolic functions which uncover underlying non-linear relationships between features of a large dataset. Our proposed GP-FRM method used a variation on Pearson's correlation with a speciation-based genetic programming algorithm to automatically produce a set of distinct and meaningful feature relationships. Empirical testing across a range of real-world datasets demonstrated the ability of GP-FRM to find very strong relationships which used a small number of features, aided by the use of parsimony pressure as a secondary objective. Further analysis reinforced these findings and demonstrated how the learned relationships could be used in practice.

Future work will primarily focus on improving GP-FRM further through the use of more sophisticated parsimony pressure methods; development of approaches to minimise the number of distinct features used in a given species; and further refinements to the fitness function to better measure the interpretability and meaningfulness of feature relationships. Employing measures such as the Shapley value \cite{roth1988shapley} or the Vapnik–Chervonenkis dimension \cite{vapnik2015uniform} could give better measures of tree complexity than a simple count of nodes.

\bibliographystyle{splncs03_unsrt}
\bibliography{biblo_august} 

\end{document}